\documentclass[runningheads]{llncs}

\usepackage{researchpack}
\usepackage{xcolor}
\usepackage{xspace}
\usepackage{hyperref}
\usepackage{subfig}
\usepackage[super]{nth}
\usepackage{placeins}

\usepackage{graphicx}
\title{
    Toward Machine-Guided, Human-Initiated\\Explanatory Interactive Learning
    \thanks{This work has received funding from the European Research Council
    (ERC) under the European Union’s Horizon 2020 research and innovation
    programme (grant agreement No.  [694980] SYNTH: Synthesising Inductive Data
    Models).  The research of ST has received funding from the ``DELPhi - DiscovEring Life Patterns'' project funded by the MIUR Progetti di Ricerca di Rilevante Interesse Nazionale (PRIN) 2017 -- DD n. 1062 del 31.05.2019.}
}

\author{
    Teodora Popordanoska\inst{1} \and
    Mohit Kumar\inst{1} \and
    Stefano Teso\inst{2}
}
\authorrunning{T. Popordanoska et al.}

\institute{
    $^1$ KU Leuven, Belgium\\
    \email{teodora.popordanoska@students.kuleuven.be},
    \email{mohit.kumar@cs.kuleuven.be}\\
    $^2$ University of Trento, Italy\\
    \email{stefano.teso@unitn.it}
}

\begin{document}
\maketitle

\begin{abstract}

    Recent work has demonstrated the promise of combining local explanations
    with active learning for understanding and supervising black-box models.
    Here we show that, under specific conditions, these algorithms may
    misrepresent the quality of the model being learned.
    The reason is that the machine illustrates its beliefs by predicting and
    explaining the labels of the query instances:  if the machine is unaware of
    its own mistakes, it may end up choosing queries on which it performs
    artificially well.
    This biases the ``narrative'' presented by the machine to the user.
    We address this \emph{narrative bias} by introducing explanatory guided
    learning, a novel interactive learning strategy in which:
    i)~the supervisor is in charge of choosing the query instances, while
    ii)~the machine uses global explanations to illustrate its overall behavior
    and to guide the supervisor toward choosing challenging, informative
    instances.
    This strategy retains the key advantages of explanatory interaction while
    avoiding narrative bias and compares favorably to active learning in terms
    of sample complexity.
    An initial empirical evaluation with a clustering-based prototype
    highlights the promise of our approach.


\end{abstract}

\section{Introduction}

The increasing ubiquity and integration of sophisticated machine learning into
our lives calls for strategies to justifiably establish or reject trust into
models learned from data~\cite{rahwan2019machine}.  Explanatory interactive
learning~\cite{teso2019explanatory,schramowski2020right} aims to achieve this
by combining interactive learning, which enables users to build expectations
through continued interaction, with computational explanations, which
illustrate the model's inner logic in an interpretable manner.

Compatibly with this observation, explanatory \emph{active} learning (XAL)
addresses classification tasks by combining active learning with local
explanations~\cite{teso2019explanatory}.  During learning, the machine selects
unlabeled instances (e.g., documents or images) and asks an annotator to label
them.  At the same time, the machine supplies \emph{predictions} for the query
instances and \emph{local explanations} for these predictions to the annotator.
The local explanations unpack the reasons behind the prediction in terms of,
e.g., feature relevance~\cite{guidotti2018survey}.  The supervisor can also
correct the local explanations by highlighting, e.g., confounding features that the machine is wrongly relying on.  A recent study on plant phenotyping data has shown that explanatory interaction helps human supervisors to acquire classifiers that are
``right for the right reasons''~\cite{ross2017right} and to correctly assign trust into them~\cite{schramowski2020right}.

Despite these results, in some situations the \emph{narrative} presented by XAL
may misrepresent the actual performance of the classifier.  The issue is that,
the narrative is focused on the query instances and the machine may fail to choose
instances that capture its own flaws.  This occurs, for instance, when the
classifier is affected by unknown unknowns, i.e., (regions of) high-confidence
mistakes~\cite{dietterich2017steps,attenberg2015beat}.  This leads to a form of
\emph{narrative bias}.

We tackle this issue by introducing explanatory guided learning (XGL), a novel
form of human-initiated interactive learning that relies on \emph{global}
explanations~\cite{andrews1995survey,guidotti2018survey}, which summarize the
whole predictor using an interpretable surrogate, e.g., clusters or rules.
Crucially, the supervisor, rather than the machine, is responsible for choosing
the query instances.  We argue that global explanations bring two key benefits.
First, they convey less biased expectations of the predictor's behavior, thus
making it possible to avoid narrative bias.  Second, they support
human-initiated query selection by guiding the annotator towards discovering
informative, problematic instances.  This novel form of \emph{human-initiated},
\emph{machine-guided} interaction retains most benefits of explanatory
interaction, including facilitating the acquisition of high-quality
classifiers.
We present an initial implementation of explanatory guided learning that uses
clustering techniques to produce data-driven global explanations, and evaluate
it empirically on a synthetic data set.  Our initial results support the idea
that explanatory guided learning helps supervisors to identify useful examples even in the
presence of unknown unknowns and sub-optimal decision making.

Summarizing, we:
1) Identify the issue of narrative bias in explanatory active learning;
2) Introduce explanatory guided learning, which avoids narrative bias by
combining human-initiated interactive learning with machine guidance in the
form of global explanations;
3) Develop a prototype implementation and present an initial empirical
evaluation on a synthetic data set.

\section{Problem Statement}

We are concerned with learning a classifier $f: \calX \to \calY$ from examples
$\calL = \{ (x_i, y_i) : i = 1, \ldots, n \} \subseteq \calX \times \calY$.
Here, $\calX$ is the space of inputs and $\calY = \{1, \ldots, c\}$ are the
labels.  The classifier is assumed to be black-box, e.g., a deep neural network
or a kernel machine.  Extra examples can be acquired by interacting with a
human supervisor.
Two requirements are put into place:
\begin{enumerate}

    \item Training data is initially scarce and obtaining more from the
        supervisor comes at a cost.  Hence, a good classifier $f$ should be
        identified using few, well-chosen queries.

    \item The supervisor should be able to tell whether $f$ can be trusted as
        objectively as possible.  The machine must supply information for this
        purpose.

\end{enumerate}
The last requirement is not easy to formalize.  Intuitively, it means that the
machine should output performance statistics, predictions, explanations,
proofs, plots, or any other kind of interpretable information necessary for the
supervisor to establish whether $f$ is trustworthy.  Clearly, providing
persuasive information that misrepresents the quality of the model is in
contrast with this requirement.

\section{Preliminaries}

The first requirement is satisfied by standard techniques like active learning
(AL)~\cite{settles2012active}.
To recap, in AL it is assumed that the machine has access to a large pool of
unlabeled instances $\calU \subseteq \calX$.  During learning, the machine
picks query instances from $\calU$, asks the supervisor to label them, and uses
the feedback to update the classifier.  The queries are chosen by maximizing
their estimated informativeness, usually defined in terms of how uncertain the
model is about their label and how well they capture the data
distribution~\cite{shui2019deep}.
The AL interaction protocol however is completely opaque and thus fails the
second requirement~\cite{teso2019explanatory}.

\begin{algorithm}[tb]

    \begin{algorithmic}[1]
        \State initialize $f$
        \Repeat
            \State select $x \in \calU$ by maximizing informativeness w.r.t. $f$
            \State present $x$, prediction $\hat{y} = f(x)$, and local explanation $z$ to the user
            \State receive ground-truth label $y$ and correction $\bar{z}$
            \State convert $\bar{z}$ to examples (see~\cite{teso2019explanatory})
            \State update $\calL$ and $\calU$, retrain $f$
        \Until{query budget exhausted or $f$ good enough}
        \State $\textbf{return} f$
    \end{algorithmic}

    \caption{\label{alg:xal}  Pseudo-code of explanatory active
    learning~\cite{teso2019explanatory,schramowski2020right}.}

\end{algorithm}

Explainable active learning (XAL) tackles this issue by supplying the user with
information about the model being
learned~\cite{teso2019explanatory,schramowski2020right}.
The learning loop (listed in Algorithm~\ref{alg:xal}) is similar to active
learning, except that, after choosing a query point $x$, the machine also
predicts its label $\hat{y} = f(x)$ and explains the prediction using a
\emph{local explanation} $z$.
Local explanations are a building block of
explainability~\cite{guidotti2018survey}:  they illustrate the logic behind
individual predictions in terms of visual artifacts (e.g., saliency maps) that
highlight which features are most responsible for the prediction.
The query $x$, prediction $\hat{y}$, and explanation $z$ are then supplied to
the supervisor.
Over time, this gives rise to a ``narrative'' that allows the supervisor to
monitor the beliefs acquired by the machine and its improvement or lack
thereof~\cite{teso2019explanatory,schramowski2020right}.
The supervisor is allowed to provide a corrected local explanations $\bar{z}$
by identifying, e.g., irrelevant features that appear as relevant in $z$.  The
corrections are translated into examples~\cite{teso2019explanatory} or gradient
constraints~\cite{schramowski2020right} and used as additional supervision.
This allows to directly teach the machine not to rely on, e.g., confounders.

Experiments with domain experts have shown that explanatory active learning
enables users to identify bugs in the model and to steer it away from wrong
concepts~\cite{schramowski2020right}.  XAL has also shown potential for
learning deep neural nets~\cite{teso2019toward}.

\subsection{Narrative Bias}

It was shown that narratives produced by XAL can work well in
practice~\cite{teso2019explanatory,schramowski2020right}.  The question is:
\emph{do such narratives always help?}

\begin{figure}[t]
    \centering
    \begin{tabular}{ccc}
        \includegraphics[width=0.3\linewidth]{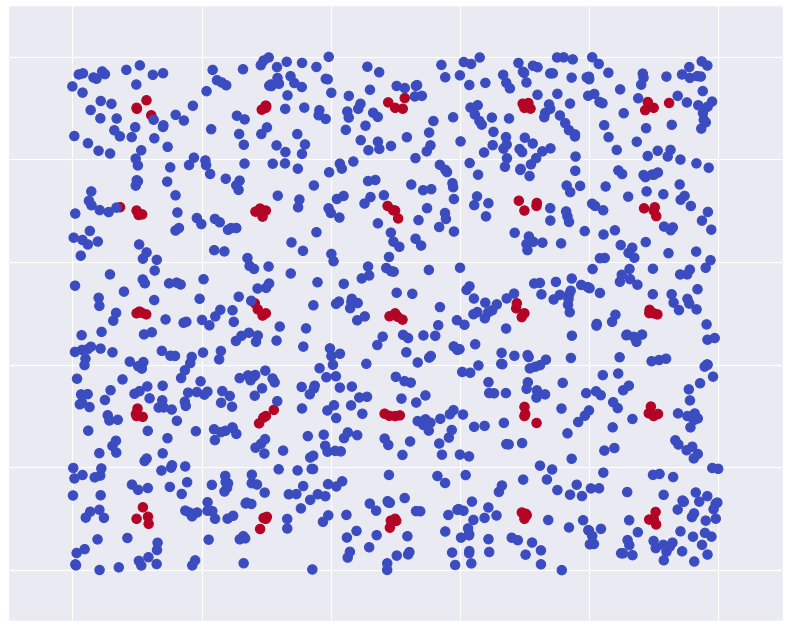} &
        \includegraphics[width=0.3\linewidth]{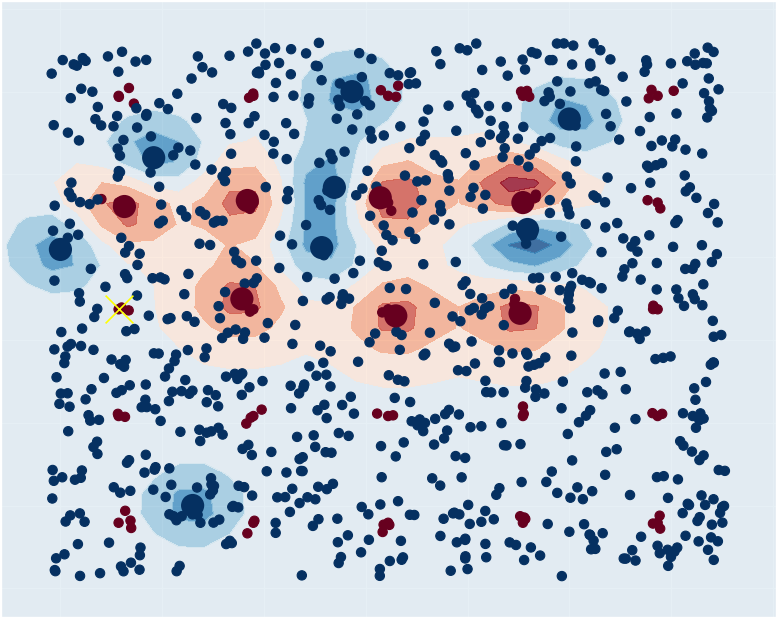} &
        \includegraphics[width=0.3\linewidth]{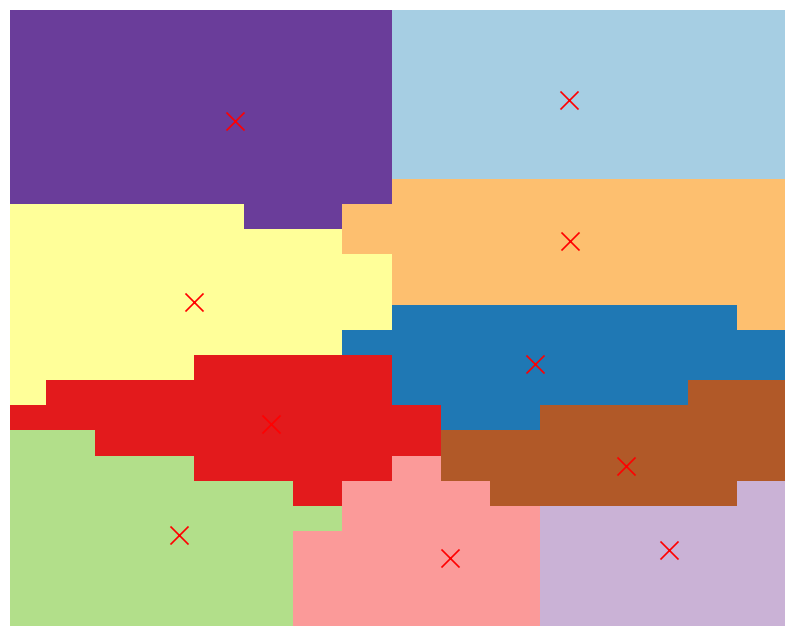}
    \end{tabular}
    \caption{Left: synthetic data set.
    Middle: predictor with unknown unknowns.
    Right: example clustering-based global explanation;  crosses are medioids.}
    \label{fig:example}
\end{figure}

The answer is no.  Narratives focused on individual query instances may
over-sell the predictor.
Consider the data set in Figure~\ref{fig:example}, left.  The red points belong
to $25$ red clusters arranged on a regular grid, while the blue ones are
distributed uniformly everywhere else.  The decision surface of an SVM with a
Gaussian kernel trained on $\approx 10$ examples is shown in the middle.
Whereas the red clusters covered by the training set are recognized as such,
the SVM is completely unaware of the other clusters. They are \emph{unknown
unknowns}~\cite{dietterich2017steps}.  What happens then is that AL sampling
strategies would choose uncertain points around the known red clusters.  At
some point, the SVM would learn the known region and thus perform well on the
query instances -- in terms of both predictions and explanations.  Therefore,
the user might get the wrong impression that the model works well
everywhere\footnote{Using other query strategies would not solve the problem
(e.g., density-based strategies would fail as the data has no density lumps),
cf.~\cite{attenberg2010label}.}.  The unknown red clusters are however highly
representative of the model's performance and should not be ignored by the
narrative.

This example shows also that unknown unknowns prevent the machine from choosing
truly informative queries.  Given that unknown unknowns occur often under class
unbalance~\cite{attenberg2010label}, sampling bias~\cite{attenberg2015beat},
and concept drift~\cite{gama2014survey}, both this and narrative bias are serious
issues in practice.

\section{Explanatory Guided Learning}

In order to tackle narrative bias, we consider a very different setup.  The
idea is that, if the supervisor could see the whole decision surface of the
predictor and were able to understand it, she could spot regions where the
predictor misbehaves and select informative supervision from these regions.
This form of human-initiated interactive learning~\cite{attenberg2010label}
would be very strong against narrative bias.
Of course, this setup is not realistic:  the decision surface of most
predictors is complex and hard to visualize, let alone validate and search
instances with.

We propose to make this strategy practical using \emph{global explanations}.
While local explanations target individual predictions, global explanations
illustrate the logic of the whole
model~\cite{andrews1995survey,guidotti2018survey}.
We restrict our attention to global explanations that
summarize~\cite{bucilua2006model} the target classifier using an interpretable
surrogate\footnote{Other kinds of global explanations, such as those based on
feature dependencies or shape
constraints~\cite{henelius2014peek,tan2018learning}, are not considered.}.

Given a classifier $f\!: \calX \to \calY$, a global explanation is a classifier
$g\!: \calX \to \calY$ that approximates $f$ and is taken from a suitable
family of interpretable predictors, like (shallow) decision
trees~\cite{craven1996extracting,krishnan1999extracting,boz2002extracting,tan2016tree,bastani2017interpreting,yang2018global}
or (simple)
rules~\cite{nunez2002rule,johansson2004accuracy,barakat2010rule,augasta2012reverse}.
Usually, $g$ is by obtained by sampling a large enough set of instances
$\{x_1', \ldots, x_s'\} \subseteq \calX$ and then solving $\min_g \sum_{k=1}^s
\ell(f(x_k'), g(x_k'))$, where $\ell(y,y')$ is some loss function.  It is
common to sample instances close to the data manifold, so to encourage the
surrogate to mimic target predictor on the bulk of the
distribution~\cite{bucilua2006model}.  For simplicity, in our experiments we
employ clustering-based explanations obtained by fitting $k$ clusters to the
data, in which the label (known for $\calL$ and predicted for $\calU$) is
treated as a feature, see Figure~\ref{fig:example} (right) for an example.

A global explanation $g$ is presented as a visual or textual
artifact~\cite{guidotti2018survey}.  In our case, a clustering-based
explanation consists of a set of $k$ clusters, each associated to its predicted
(majority) label and a textual description like ``feature $1$ is larger than
$10$ and feature $4$ is less than $35$ and \ldots''.  Thanks to their
interpretability, global explanations are a natural device for helping
supervisors to spot mistakes and also to select impactful examples, as it makes
it possible for users to formulate counter-examples to, e.g., clearly wrong
rules or clusters.  

\begin{algorithm}[tb]

    \begin{algorithmic}[1]
        \State fit classifier $f$ on $\calL$, compute global explanation $g \approx f$ \label{line:initialize}
        \Repeat
            \State supply $g$ to user and ask for an example \label{line:query}
            \State receive $(x, y)$ from user and add it to $\calL$ \label{line:receive}
            \State update $f$ using $\calL$ \label{line:updatef}
            \State update $g \approx f$ \label{line:updateg}
        \Until{query budget exhausted or $f$ good enough}
        \State $\textbf{return} f$
    \end{algorithmic}

    \caption{\label{alg:xgl}  Explanatory guided learning.  $\calL$ is the
    training set.}

\end{algorithm}

We call the combination of global explanations and human-initiated interactive
learning \emph{explanatory guided learning}.  The pseudo-code is listed in
Algorithm~\ref{alg:xgl}.  The learning loop is straightforward.  Initially a
classifier $f$ is learned on the initial training set
$\calL$ and a global explanation $g \approx f$ is computed (line
\ref{line:initialize}).  Then the interaction loop beings.  In each iteration,
the machine presents the global explanation $g$ to the supervisor and asks for
a high-loss example (lines \ref{line:query}--\ref{line:receive}).  This is
discussed more in detail below.  Upon receiving new supervision, the machine
updates the training set, the predictor $f$, and the global explanation $g$
(lines \ref{line:receive}--\ref{line:updateg}).  The loop repeats until the
classifier is deemed good enough or the labeling budget is exhausted.

\subsection{Discussion}

A key advantage of XGL is that it is -- by design -- immune to the form of
narrative bias discussed above.  A second key advantage is that it
enables supervisors to provide examples tailored to the model at hand.  This is
critical in the presence of unknown unknowns and in other cases in which
machine-initiated interactive learning fails~\cite{attenberg2010label}.  Our
preliminary experimental results are consistent with this observation.
Notice that simply combining global explanations with machine-guided learning
would \emph{not} achieve the same effect, as the learning loop would entirely
depend on possibly uninformative queries selected by the machine.  Similarly, using a held-out validation set to monitor the model behavior would not capture the same information conveyed by global explanations.
Another advantage is that global explanations offer support for protocols in
which supervisors select entire batches of data rather than individual
examples, as usually done in active learning of deep models~\cite{gal2017deep}.

Naturally, shifting the responsibility of choosing instances from the machine
to the user may introduce other forms of bias.  For instance, the explanation
may be too rough an approximation of the target model or the supervisor may
misinterpret the explanation.  These two issues, however, are not exclusive to
XGL:  local explanations can be unfaithful~\cite{teso2019toward} and
annotator performance can be poor even in AL~\cite{zeni2019fixing}.

The main downsides of global explanations over local explanations are their
added cognitive and computational costs.  Despite this issue, we argue that
global explanations are necessary to avoid narrative bias, especially in
high-risk applications where the cost of deploying misbehaving models is
significant.
Moreover, the computational cost can be amortized over time by making use of incremental
learning techniques.  The cognitive cost can also be reduced and diluted
over time, for instance by restricting the global explanations to regions that
the user cares about.  Another possibility is to employ a mixed-initiative
schema that interleaves machine- and human-initiated interactive learning.
This would make global explanations less frequent while keeping the benefits of
XGL.  The question becomes when to show the global explanations.  One
possibility is to program the machine to warn the user whenever the feedback
has little impact on the model, indicating that either the query selection algorithm is ``stuck'' or that the supervisor's understanding of the machine is misaligned.

We remark that our clustering-based implementation is meant as a prototype.
More refined implementations would use global explanations based on trees or
rules~\cite{guidotti2018survey} and provide the user with interfaces and search tools to explore the space using the global explanation.  For instance, the interface could build on the one designed for guided learning~\cite{attenberg2010label}, a form of human-initiated interactive learning, by supplementing it with explanations.

\section{Experiments}

We study the following research questions:
\begin{itemize}

    \item[\textbf{Q1}] Is XGL less susceptible to narrative bias than machine-initiated alternatives?

    \item[\textbf{Q2}] Is XGL competitive with active learning and guided learning in terms of sample complexity and model quality?

    \item[\textbf{Q3}] How does the annotator's understanding of the global explanation affect the performance of XGL?

\end{itemize}
\textbf{Experimental Setup:}
To answer these questions, we ran our clustering-based prototype on a synthetic classification task and compared it with several alternatives.  The data set is illustrated in Figure~\ref{fig:example} (left).  The data consists of an unbalanced collection of $941$ blue and $100$ red ($100$ points) bi-dimensional points, with a class ratio of about $10:1$.  The red points were sampled at random from 25 Gaussian clusters distributed on a five by five grid.  The blue points were sampled uniformly from outside the red clusters with little or no overlap.  All results are $10$-fold cross-validated using stratification. For each fold, the training set initially includes five examples, at least two per class. The implementation and experiments can be found at \url{https://bit.ly/32SJUhB}.

\textbf{Human-machine interaction:} 
The global explanation presented to the human supervisor acts as a summary of the model's behavior on different regions of the problem space. In our experiments, the summary is constructed from clusters obtained from the data using $k$-medoids. For the synthetic data set, we extract 10 clusters. In general, the number of clusters can either be determined in advance by the system designer, or it can be dynamically adapted based on the desired precision of the explanation. 
Each cluster is represented by a prototype - an exemplar case that serves to approximate the behavior of the model on that region as a whole. In addition to the prototypes, the regions can be further described, e.g., by utilizing interpretable decision trees~\cite{bastani2017interpreting,craven1996extracting,yang2018global} and extracting a rule list as a textual description.
Upon inspection of the prototypes, their corresponding predicted labels and the description of the clusters, the human is expected to identify the ones with high loss and provide instances that correct the model's beliefs. In other words, the search strategy performed by the user has a hierarchical structure: first she chooses a region where the model misbehaves and then she looks for an instance within that region based on some criterion.

\textbf{User simulation:} 
A helpful and cooperative teacher is simulated with a simple model that attempts to capture the various levels of knowledge and attention of real users. In the optimistic case, the simulated user is able to consistently detect a region of weakness for the learner. In the experiments, the cluster having the most wrongly classified points is  regarded  as  the  weakest  area  for  the  learner. In the worst case, the teacher selects an instance from a randomly chosen cluster. Within the chosen cluster, the simulated user selects a wrongly classified point closest to its prototype. 

\textbf{Baselines:}
We compare the performance of our method against the following baselines: 
1) \textbf{Guided learning:} This strategy is simulated by class-conditional random sampling, i.e., the user interchangeably chooses instances from each class in a balanced proportion. 
2) \textbf{Active learning:} Following the most popular AL strategy -- uncertainty sampling -- the instances are chosen based on the uncertainty of the classifier in their label. 
3) \textbf{Random sampling:} The instances to be labeled are uniformly sampled from the unlabeled pool. This simple baseline is surprisingly hard to beat in practice.
4) \textbf{Passive learning:} The classifier is trained on the whole data. This baseline indicates how fast the methods will converge to the performance of the fully trained model.

\subsection{Q1: Is XGL less susceptible to narrative bias?}

\begin{figure}[tb]
    \centering
    \captionsetup[subfigure]{justification=centering}
    \subfloat[\nth{10} iteration]
        {
        \includegraphics[width=.3\linewidth]{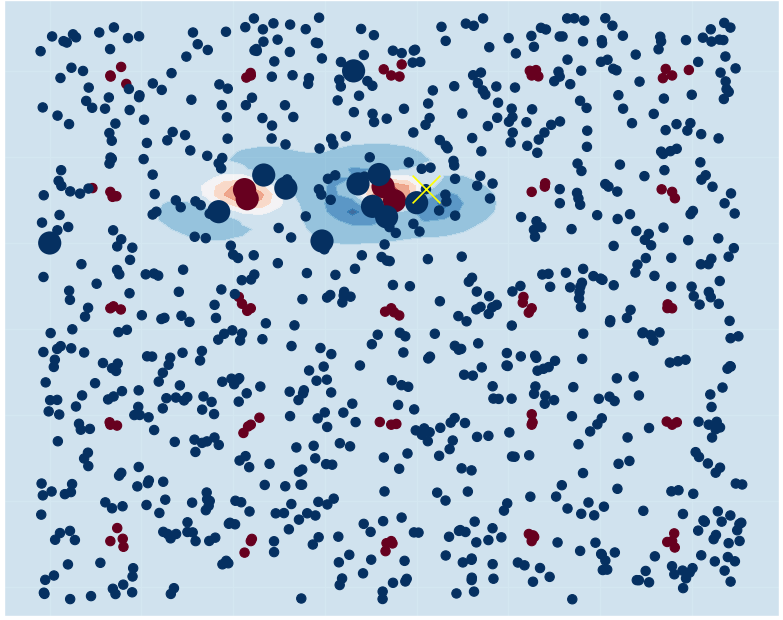}
        } \hfill
    \subfloat[\nth{70} iteration]
        {
        \includegraphics[width=.3\linewidth]{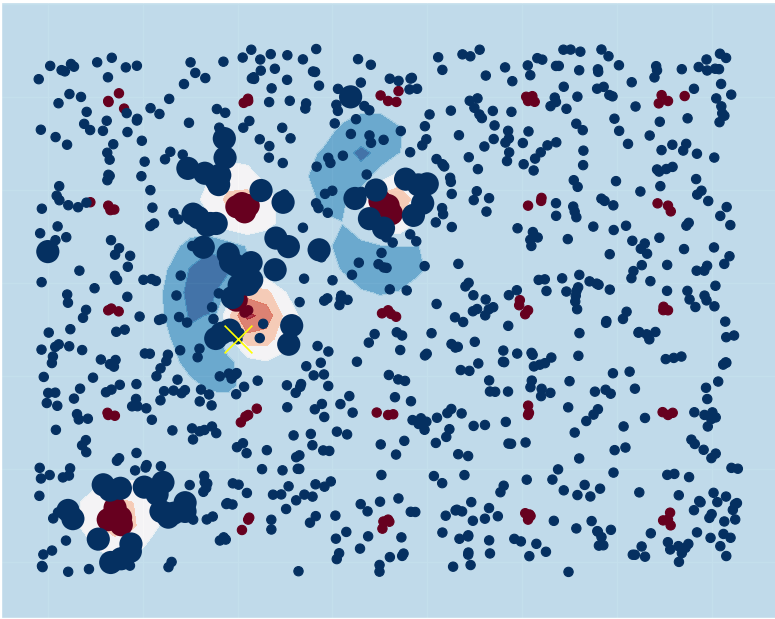}
        } \hfill
    \subfloat[\nth{140} iteration]
        {
        \includegraphics[width=.3\linewidth]{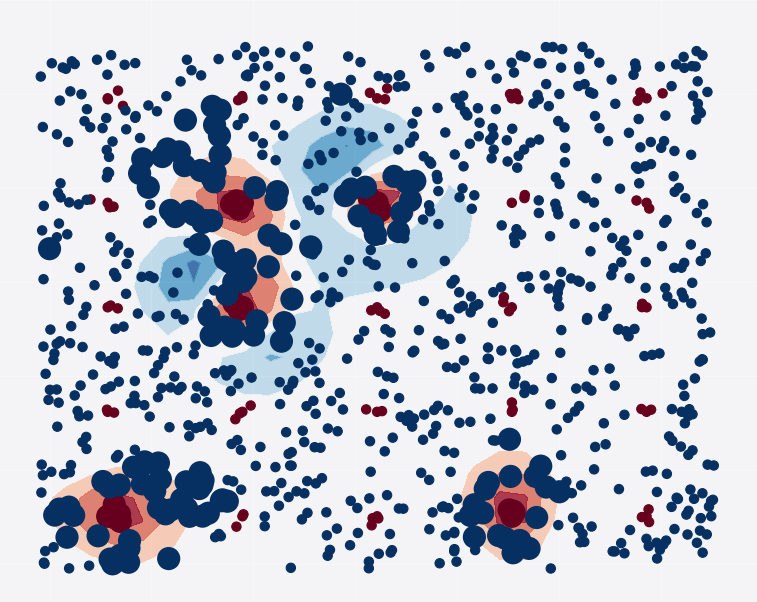}
        } \hfill
        \par 
    \subfloat[\nth{10} iteration]
        {
        \includegraphics[width=.3\linewidth]{figures/experiments/egl_10it.png}
        } \hfill
    \subfloat[\nth{70} iteration]
        {
        \includegraphics[width=.3\linewidth]{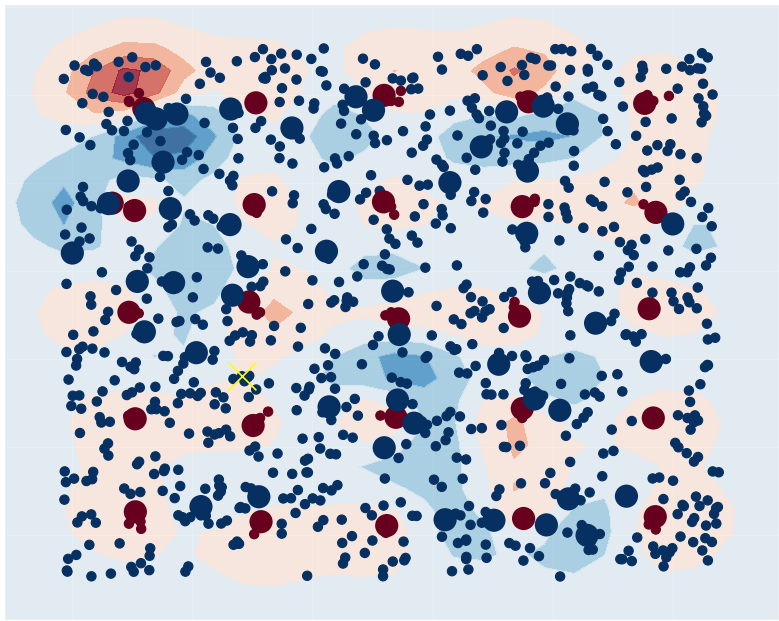}
        } \hfill
    \subfloat[\nth{140} iteration]
        {
        \includegraphics[width=.3\linewidth]{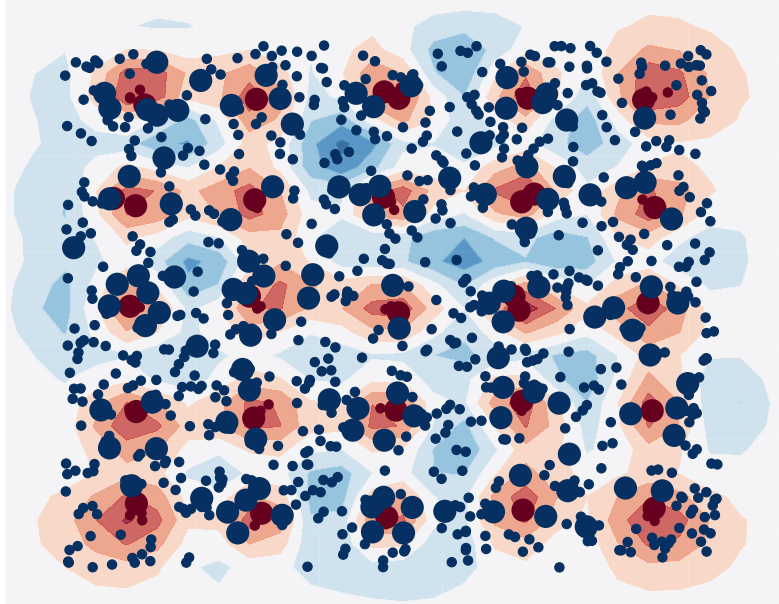}
        } \hfill
        \par 
    \caption{Results for AL vs. XGL. First row: Instances chosen by the machine using uncertainty sampling misrepresent the model's behavior in the presence of unknown unknowns. The model is oblivious to the existence of the clusters outside of the already found ones, i.e., the machine is unable to detect its own misbehavior. Second row: XGL combats narrative bias by injecting global explanations that enable the human supervisor to identify the flaws of the model and choose informative, non-redundant examples accordingly. }
    \label{fig:al_vs_egl}
\end{figure}

The first experiment investigates the methods' ability to handle unknown unknowns, i.e., red clusters that the model has not yet identified. Figure~\ref{fig:al_vs_egl} shows a comparison of the instances selected with AL using uncertainty sampling (top row) and XGL (bottom row). The exploitative nature of uncertainty sampling leads the model to select instances around the presumed decision surface of the already found red clusters, thus wasting the querying budget on redundant instances. The narrative that XAL would create based on this choice of points is not representative of the generalization (in)ability of the model. In other words, there exist many regions in the data that are not explored, because this strategy becomes locked in a flawed hypothesis of where the decision boundary is and fails to explore the space.
  
This experiment showcases that explanatory strategies rooted in AL would misrepresent the true performance of the model in the presence of unknown unknowns. Therefore, the supervisor would be wrongly persuaded to trust it. In contrast, the explanatory component of XGL enables the user to understand the beliefs of the model being learned, while the human-initiated interaction allows the supervisor to appropriately act upon the observed flaws of the model.  These preliminary results show that our prototype can be very helpful in detecting areas of wrong uncertainty and avoiding narrative bias. 

\subsection{Q2: Is XGL competitive with active learning and guided learning in terms of sample complexity and model quality?} 

\begin{figure}[!htbp]
    \centering
        \includegraphics[width=0.49\textwidth]{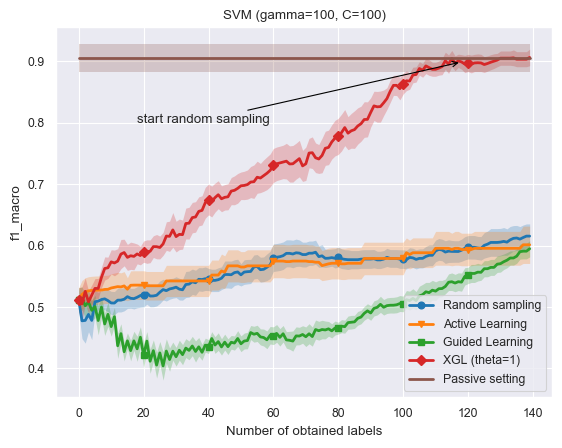}
        \includegraphics[width=0.49\textwidth]{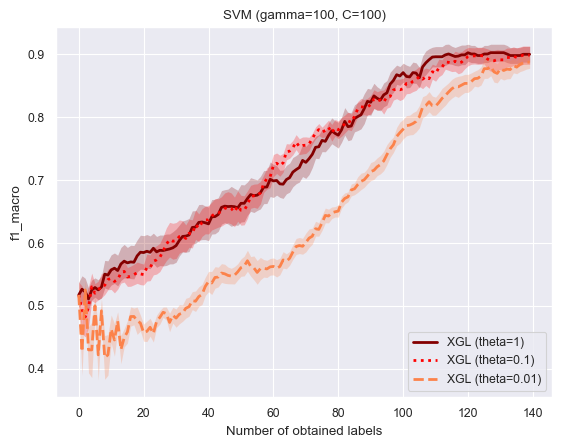} 
    \caption{Left: F1 score on the synthetic data set with SVM ($\gamma$=100, $C$=100).  Our prototype surpasses the alternatives by a large margin. Right: The performance of XGL for simulated users with varying parameter $\theta$ in a softmax function over the number of misclassified points in every cluster.}
    \label{fig:methods_and_thetas}
\end{figure}

To address this question, we compare the F1 score versus the number of labeled examples, shown in Figure~\ref{fig:methods_and_thetas} (left). The performance is calculated using predictions on a separate test set. In every iteration, one instance is selected to be labeled using the strategies of interest. The model is retrained and the accuracy is reported.

To ensure that all methods have received the same amount of supervision, after the pool of wrongly classified points for XGL is exhausted, the simulation continues to sample random points from the unlabeled data. The iteration when the switch happens is depicted with the arrow on the plot. Notice that by that iteration, the model already achieves the same performance as the fully trained model.

\begin{figure}[!htbp]
    \centering
        \includegraphics[width=0.45\textwidth]{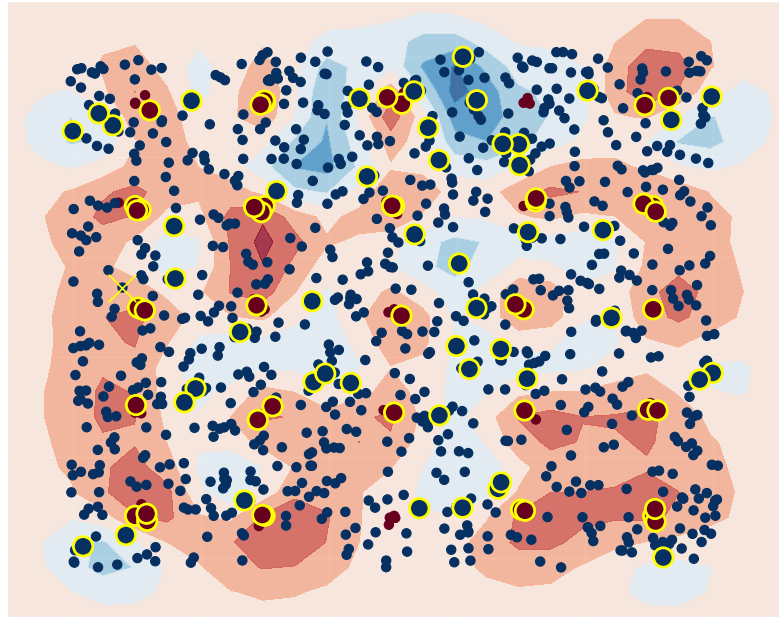} 
        \includegraphics[width=0.45\textwidth]{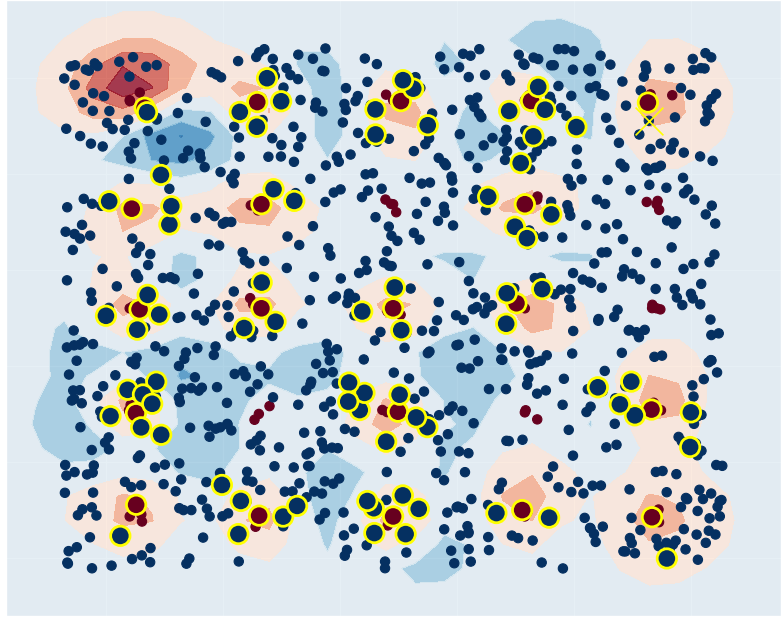}
    \caption{Decision surfaces in the \nth{90} iteration. Left: GL. No explanations are shown to the supervisor. Consequently, a lot of redundant points are selected (red points from already found red clusters). Right: XGL. The supervisor is presented with clustering-based global explanation. The chosen instances are balanced between refining the decision boundary and exploring new red clusters.}
    \label{fig:gl_vs_egl}
\end{figure}

In the initial stages of learning, the classifier is oblivious to the existence of red clusters outside of the assumed decision boundary around the labeled data points. In these conditions, the query selection with uncertainty sampling, as a representative for \textbf{active learning}, triggers a vicious cycle of selecting instances that add little information for the update of the classifier, which in turn leads to even more uninformative instances chosen in the next iteration because of the poor quality model. Consequently, in the given budget of queries, the model discovers only a fraction of the red clusters, resulting in a poor overall performance. Active learning and random sampling only rarely query instances from the red class, which is the reason for their slow progress shown in  Figure~\ref{fig:methods_and_thetas} (left). On the other hand, using \textbf{guided learning}, in every iteration the model interchangeably observes instances from the two classes. The lack of sufficient blue points queried to refine the already found red clusters, leads the model to create a decision boundary as illustrated in  Figure~\ref{fig:gl_vs_egl} (left), where many of the blue points will falsely be classified as red. 

The decision surfaces of the classifier trained on the points selected by the strategies based on guided learning are shown in Figure \ref{fig:gl_vs_egl}. Comparing the chosen instances (circled in yellow) with each of these methods, it is evident that by using uninformed guided learning the supervisor is likely to present the learner instances from regions where it is already performing well. This observation, once again, emphasizes the importance of global explanations for enabling the user to provide non-redundant supervision, which ultimately results in more efficient learning in terms of sample complexity.

In summary, the overall trend is consistent with our intuition: using XGL we can train a significantly better classifier at low labeling costs, compared to the alternatives. 

\subsection{Q3: How does the annotator's understanding of the global explanation affect the performance of XGL?}

Needless to say, when the supervisor has a central role in the model's learning process, understanding the explanation and taking proper actions becomes crucial. However, in realistic scenarios, human annotators can be imprecise and inconsistent in identifying regions with high loss. 
To account for these situations in our preliminary experiments, we simulate different users with a softmax function, parametrized by $\theta$, over the number of misclassified points in every cluster. 
Let $m_i$ denote the number of mislabeled points in cluster $i$. The probability of the user choosing the cluster $i$ is given by:
\begin{equation}
  \label{eq:softmax}
  P(\text{choose cluster $i$}) = \frac{\exp(\theta m_i)}{\sum_{j=1}^k \exp(\theta m_j)},
\end{equation}
where $\theta$ is a parameter that serves to simulate the different users. Larger $\theta$ simulates a supervisor who identifies the weakest region for the classifier and chooses to label data points from it. In the worst case, the annotator does not understand the presented explanation and chooses a cluster at random, which is simulated with smaller $\theta$. The results obtained for different values of $\theta$ are presented in Figure~\ref{fig:methods_and_thetas} (right). It can be observed that significant improvements can be gained for reasonable choices of clusters to select instances from, as simulated with $\theta=1$ and $\theta=0.1$.

\section{Related Work}

The link between interactive learning, explainability, and trust is largely
unexplored.  Our work is rooted in explanatory interactive
learning~\cite{teso2019explanatory,schramowski2020right} (see
also~\cite{sen2018supervising} and~\cite{phillips2018interpretable}), of which
XGL and XAL are instantiations.

There is little work on human-initiated interactive learning.  XGL is an
extension of guided learning~\cite{attenberg2010label,simard2014ice}, in which
search queries are used to combat label skew in classification tasks.  We
deepen these insights and show that global explanations combined with
human-initiated interaction can be a powerful tool for handling  unknown unknowns~\cite{dietterich2017steps}. A major
difference with our work is that guided learning is entirely black-box:  the
annotator is asked to provide examples of a target class but receives no
guidance (besides a description of that class).  Since the supervisor has no
clue of what the model looks like, this makes it difficult for her to establish
or reject trust and also to provide useful (e.g., non-redundant) examples.  In
contrast, XGL relies on global explanations to guide the user.  Let us note
that guided learning compares favorably to pure active learning in terms of
sample complexity~\cite{beygelzimer2016search}.  The
idea of asking supervisors to identify machine mistakes has recently been
explored in~\cite{attenberg2015beat,vandenhof2019contradict}, but the
relationship to global explanations as machine guidance is ignored.  

Our observations are consistent with recent work in interactive machine
teaching.  Machine teaching is the problem of selecting a minimal set of
examples (a ``teaching set'') able of conveying a target hypothesis to a
learner~\cite{zhu2015machine}.  The focus is primarily theoretical, so it is
typically assumed that the teacher (who designs the teaching set) is a
computational oracle with unbounded computational resources and complete access
to the model and learning algorithm.  It was recently proved that oblivious
teachers unaware of the state of the model cannot perform better than random
sampling~\cite{melo2018interactive,chen2018near,dasgupta2019teaching}.  In
order to overcome this limitation, the teacher must interact with the machine,
as we do.  Existing algorithms, however, cannot be applied to human oracles or
assume that the teacher can sample the whole decision surface of the learner,
and in general ignore the issue of trust.  Our work identifies global
explanations as a practical solution to all of these issues.

Explanatory guided learning revolves around machine-provided guidance in the
form of global explanations.  This is related to, but should not be confused
with, work on user-provided guidance~\cite{kunapuli2013guiding} and
teaching guidance~\cite{cakmak2014eliciting}.  These
are orthogonal to our approach and could be fruitfully combined with it.

\section{Conclusion and Outlook}

The purpose of this paper is twofold.
On the one hand, it draws attention to the issue of narrative bias, its root
causes, and its consequences on trust in explanatory active learning.
On the other, it shows how to deal with narrative bias by combining
human-initiated interactive learning with machine guidance in the form of global explanations.
An initial empirical analysis suggest that explanatory guided learning, our
proposed method, helps the supervisor to select substantially less biased examples.
Of course, a more thorough validation on real-world data sets and more refined forms of
explanation, like rules or decision trees, is needed.  This is left for future
work.

\bibliographystyle{splncs04}
\bibliography{paper}
\end{document}